%% file: top.tex
\documentclass[10pt,twocolumn,letterpaper]{article}

\usepackage{cvpr}
\usepackage{times}
\usepackage{epsfig}
\usepackage{graphicx}
\usepackage{amsmath}
\usepackage{amssymb}
\usepackage{bm}

\usepackage{array}
\newcolumntype{?}{!{\vrule width 1pt}}
\newcolumntype{C}[1]{>{\centering\arraybackslash\hspace{0pt}}p{#1}}

\input{defs}

\usepackage[pagebackref=true,breaklinks=true,letterpaper=true,colorlinks,bookmarks=false]{hyperref}
\newtheorem{theorem}{Theorem}[section]
\newenvironment{proof}[1][Proof]{\begin{trivlist}
\item[\hskip \labelsep {\bfseries #1}]}{\end{trivlist}}
 \cvprfinalcopy 

\def\cvprPaperID{1254} 

\DeclareMathOperator*{\argmin}{\arg\!\min}

\ifcvprfinal\fi
\begin{document}

\title{Principled Parallel Mean-Field Inference for Discrete Random Fields}

\author{Pierre Baqu{\'{e}}$^1$\thanks{ \:  The authors contributed equally.}\quad\quad Timur Bagautdinov$^1$$^{*}$ \quad\quad Fran{\c{c}}ois Fleuret$^{1,2}$ \quad\quad Pascal Fua$^1$ \\
$^1$CVLab, EPFL, Lausanne, Switzerland\\
$^2$IDIAP, Martigny, Switzerland\\
{\tt\small \{firstname.lastname\}@epfl.ch}
}

\maketitle

\begin{abstract}
Mean-field variational  inference is one  of the  most popular approaches to 
inference in discrete random fields.  Standard mean-field optimization is based on
coordinate descent and in many situations can be impractical. Thus, in practice,
various parallel  techniques are used, which  either rely on \textit{ad hoc} smoothing with
heuristically set parameters, or put strong constraints on the type of models.

In this paper, we propose a novel proximal gradient-based approach to optimizing
the variational objective. It is naturally parallelizable and easy to implement. 

We  prove  its convergence, and then  demonstrate that,  in practice,  it
yields faster  convergence and often  finds better optima than  more traditional
mean-field optimization techniques. Moreover, our method is less sensitive to the choice of
parameters. 
\end{abstract}

\vspace{-0.2cm}
\section{Introduction}
\label{sec:intro}
\input{introduction}

\section{Background and Related Work}
\label{sec:related}
\input{related}

\section{Method}
\label{sec:method}
\input{method}

\section{Experimental Evaluation}
\label{sec:evaluation}
\input{evaluation}

\section{Discussion and Future Work}
\label{sec:discussion}

\input{discussion}

{\small
  \bibliographystyle{ieee}
  \bibliography{string,vision,learning,optim,refs}
}

\end{document}

%% file: defs.tex
\newcounter{nbdrafts}
\makeatletter
\newcommand{\checknbdrafts}{
\ifnum \thenbdrafts > 0
\@latex@warning@no@line{**********************************************************************}
\@latex@warning@no@line{* The document contains \thenbdrafts \space draft note(s)}
\@latex@warning@no@line{**********************************************************************}
\fi}

\makeatother

\newcommand{\comment}[1]{}

\newcommand{\bX}[0]{\mathbf{X}}
\newcommand{\bx}[0]{\mathbf{x}}
\newcommand{\bI}[0]{\mathbf{I}}
\newcommand{\bq}[0]{\mathbf{q}}
\newcommand{\bD}[0]{\mathbf{D}}
\newcommand{\bv}[0]{\mathbf{v}}
\newcommand{\bmm}[0]{\mathbf{m}}
\newcommand{\btheta}[0]{\mathbf{\theta}}

\newcommand{\mE}[0]{\mathcal{E}}
\newcommand{\mH}[0]{\mathcal{H}}
\newcommand{\mF}[0]{\mathcal{F}}

\newcommand{\mX}[0]{\mathcal{X}}
\newcommand{\mM}[0]{\mathcal{M}}

\newcommand{\KL}[0]{\text{KL}}

\newcommand{\sttt}[1]{{\footnotesize{\texttt{#1}}}}

%% file: introduction.tex

Many  Computer  Vision  problems,  ranging  from  image  segmentation  to  depth
estimation from  stereo, can be naturally  formulated in terms of  Conditional Random
Fields (CRFs). Solving  these problems then requires either  estimating the most
probable state  of the  CRF, or  the marginal  distributions over  the unobserved
variables.  Since  there  are  many such  variables,  it  is  usually
impossible  to  get  an  exact  answer,  and one  must  instead  look  for  an
approximation.

\begin{figure}[ht!]
\begin{center}
\begin{tabular}{@{}cccc}
\includegraphics[width=0.1\textwidth]{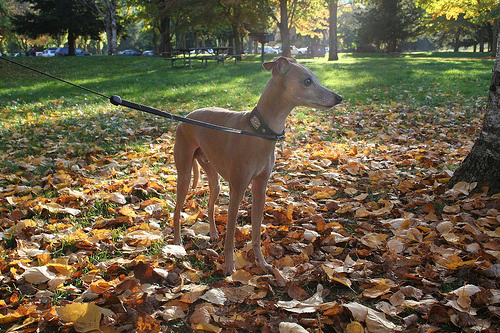} &
\includegraphics[width=0.1\textwidth]{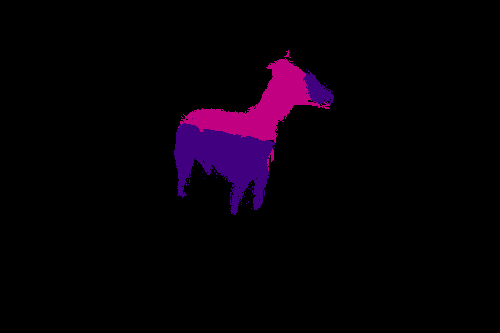} &
\includegraphics[width=0.1\textwidth]{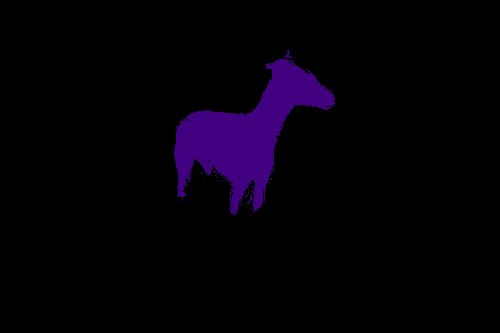} &
\includegraphics[width=0.1\textwidth]{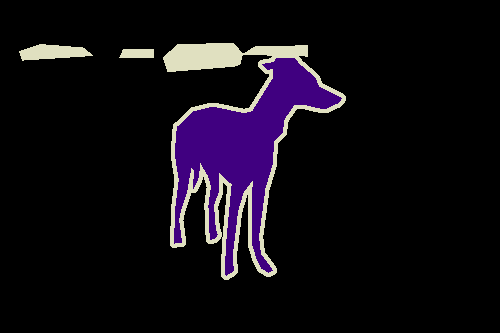} \\
\includegraphics[width=0.1\textwidth]{2007_007235_src.png} &
\includegraphics[width=0.1\textwidth]{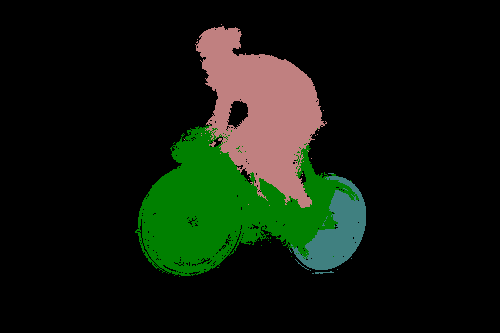} &
\includegraphics[width=0.1\textwidth]{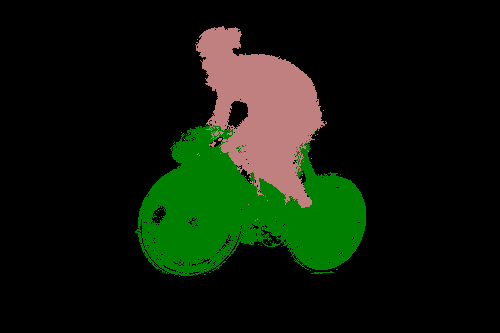} &
\includegraphics[width=0.1\textwidth]{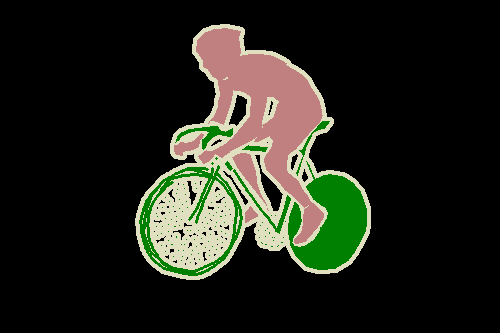} \\
\includegraphics[width=0.1\textwidth]{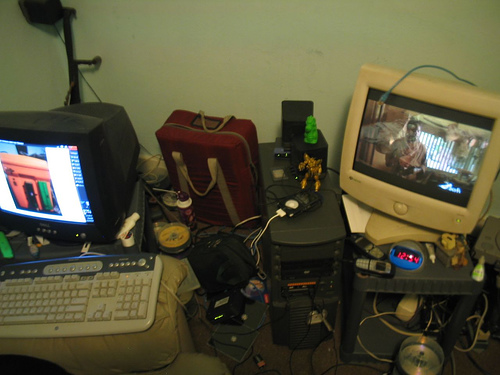} &
\includegraphics[width=0.1\textwidth]{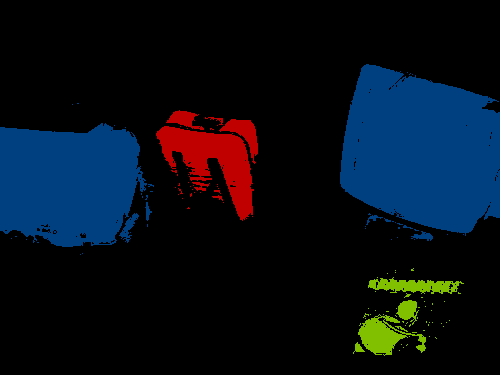} &
\includegraphics[width=0.1\textwidth]{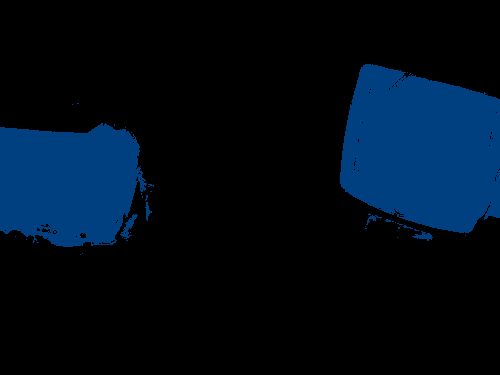} &
\includegraphics[width=0.1\textwidth]{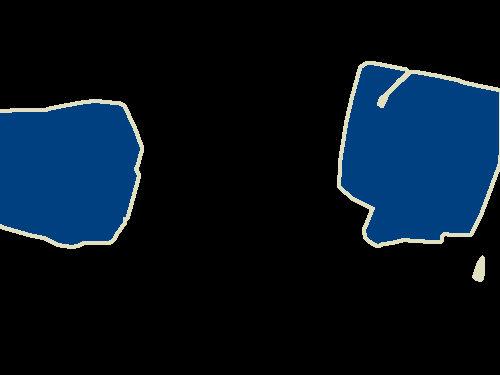} \\
\includegraphics[width=0.06\textwidth]{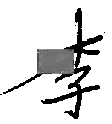} &
\includegraphics[width=0.06\textwidth]{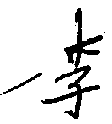} &
\includegraphics[width=0.06\textwidth]{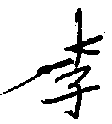} &
\includegraphics[width=0.06\textwidth]{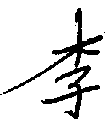} \\
\small{input} & \small{baseline} & \small{ours} & \small{ground truth} \\
\end{tabular}
\end{center}
\caption{{\bf First three rows:} VOC2012 images in which 
we  outperform a  baseline  by  adding simple  co-ocurrence  terms, 
which  our optimization scheme, unlike  earlier ones, can handle.  {\bf 
  Bottom row:} Our scheme also allows us to improve upon a baseline for 
the purpose of recovering a character from its corrupted version.}
\vspace{-0.5cm}
\label{fig:intro:teaser}
\end{figure}

Mean-field  variational   inference~\cite{Wainwright08}  is  one  of   the  most
effective  ways   to  do approximate inference and  has   become  increasingly  popular   in  our 
field~\cite{Saito12,Vineet14,Kraehenbuehl13}.  It involves introducing a variational distribution that is a
product of terms,  typically one per hidden variable. These  terms are then
estimated by minimizing the Kullback-Leibler (KL) divergence between the variational and
the true posterior.  The standard scheme is to iteratively update each factor of the
distribution        one-by-one.        This        is       guaranteed        to
converge~\cite{Bishop06,Koller09}, but is not very scalable, because all variables have to be
updated sequentially. It becomes impractical for
realistically-sized problems when there are substantial interactions between the
variables.  This can be remedied by replacing the sequential updates by parallel
ones, often at the cost of failing to converge. 

It has nonetheless recently been  shown that parallel updates
could be done in  a provably 
convergent  way for  pairwise  CRFs,  provided that  the  potentials  are
concave~\cite{Kraehenbuehl13}.  When  they are not, an {\it ad hoc} heuristic  designed to achieve
convergence, which essentially smooths steps by averaging between the next and current iterate, has
been  used over the years.  
This heuristic is mentioned  explicitly in  some works~\cite{Sun13,Campbell13,Frostig14},  or used 
implicitly  in  optimization schemes~\cite{Fleuret08a,Vineet14}  by  introducing
an additional damping parameter.

However, a  formal justification  for such smoothing is  never provided,
which we do  in this paper. More  specifically, we show that, by  damping in the
natural parameter  space instead of  the mean-parameter one, we  can reformulate
the optimization scheme  as a specific form of proximal  gradient descent.  This
yields a theoretically sound and practical  way to chose the damping parameters,
which guarantees convergence,  no matter the shape of the  potentials. When they
are   attractive,  we   show   that   our  approach   is   equivalent  to   that
of~\cite{Kraehenbuehl13}. However,  even when they  are repulsive and  can cause
the  earlier methods  to oscillate  without  ever converging,  our scheme  still
delivers  convergence.  For  example, as  shown in  Fig.~\ref{fig:intro:teaser},
this  allows us to  add co-occurrence  terms to  the model used  by a
state-of-the-art  semantic segmentation  method~\cite{Chen15b} and  improves its
results.  Furthermore,  we retain the  simplicity of the  closed-form mean-field
update rule, which is one of the key strengths of the mean-field approach.

In short, our contribution is threefold:
\vspace{-0.1cm}
\begin{itemize}
  \setlength{\itemsep}{5pt}
  \setlength{\parskip}{0pt}
  \setlength{\parsep}{0pt}
  
   \item We introduce a principled, simple, and efficient approach to performing
     parallel inference in  discrete random fields.  We  formally prove that
       it  converges and demonstrate  that  it  performs better  than
     state-of-the-art inference methods on  realistic
     Computer  Vision  tasks such  as  segmentation  and people  detection.

   \item We show that many of the earlier methods can be interpreted as variants
     of ours. However, we offer a principled way to set its metaparameters.
  
   \item  We demonstrate  how parallel mean-field  inference  in random  fields relates  to
     the gradient descent. This allows us to integrate advanced  gradient descent
     techniques, such as  momentum and  ADAM~\cite{Kingma14}, which  makes
     mean-field inference even more powerful.
\end{itemize}
To validate  our approach,  we  first  evaluate its  performance  on  a set  of
standardized benchmarks,  which include a  range of inference problems  and have
recently  been  used  to  assess inference  methods~\cite{Frostig14}.   We  then
demonstrate that  the performance improvements  we observed carry over  to three
realistic  Compute Vision  problems,  namely  Characters Inpainting,
People Detection and Semantic Segmentation.  In each case, we show that
modifying the optimization  scheme while retaining the  objective function  of
state-of-the-art   models~\cite{Fleuret08a,Nowozin11,Chen15b}  yields   improved
performance   and    addresses   the    convergence   issues    that   sometimes
arise~\cite{Vineet14}.

%% file: related.tex

In this  section, we briefly review basic Conditional Random Field  (CRF) theory and
the use of mean-field inference to solve the resulting optimization
problems. We also give a short introduction into proximal gradient descent algorithms, on which our
method is based. Note, in this work, we focus on models involving discrete random variables. 

\subsection{Conditional Random Fields}

Let $\bX = \left(X_1,  \ldots, X_N\right)$ represent hidden variables and $\bm{I}$ represent
observed variables.  For example, for  semantic segmentation, the
$X_i$s  are taken to be variables representing semantic classes of $N$ pixels,
  and  $\bm{I}$  represents the  observed  image  evidence.

A Conditional Random Field (CRF) models the relationship between  $\bX$ and $\bI$
in terms of the posterior distribution
\begin{equation}
\vspace{-0.1cm}
P(\bX \mid \bI) = \exp\left(\sum_{ c \subset\{1,\dots,N\} }\bm{\phi}_c(\bX_c \mid \bI) - \log Z
  (\bI) \right)\; ,
\label{eq:related:mrf}
\end{equation}
where $\bm{\phi}_c(.)$  are non-negative functions  known as potentials  and $\log
Z(\bI)$ is the  log-partition function. It is  a constant that we  will omit for
simplicity since  we are  mostly concerned by estimating  values of  $\bX$ that
maximize $P(\bX \mid \bI)$.

This model is often further simplified  by only considering unary and pairwise terms:
\begin{equation}
\vspace{-0.1cm}
P(\bX \mid  \bI) \propto  \exp \left( \sum_{i}  \phi_i(X_i, I_i) +  \sum_{(i,j)} \phi_{ij}(X_i,
  X_j) \right).
 \label{eq:related:pairwise-mrf}
\end{equation}

\subsection{Mean-Field Inference}

Typically, one  wants either to estimate  the posterior $P(\bX|\bI)$ or  to find
the vector  $\hat{\bX}$ that maximizes $P(\bX|\bI)$,  which is known as  the MAP
assignment. Unfortunately, even
for the  simplified formulation  of Eq.~\ref{eq:related:pairwise-mrf},  both are
intractable for realistic sizes of $\bX$. As
a  result, many  approaches  settle for  approximate  solutions.  These  include
sampling methods, such as Gibbs sampling~\cite{Gelfand90}, and deterministic ones
such as mean-field variational inference~\cite{Winn2005}, belief
propagation~\cite{Murphy99,Minka01,Kolmogorov15}, and others~\cite{Boykov01b,Gorelick2014}. 

Note that, mean-field  methods have  been  shown  to combine the advantages of good
convergence  guarantees~\cite{Bishop06}, flexibility with respect to the potential functions that
can be handled~\cite{Saito12}, and potential for parallelization~\cite{Kraehenbuehl13}. 
As a  result, they have become  very popular  in our field. Furthermore, they  have recently been
shown to yield state-of-the-art performance for several Computer Vision
tasks~\cite{Saito12,Vineet14,Chen15b,Zheng15}. 

Mean-field involves introducing a distribution $Q$ of the factorized form
\begin{equation}
\vspace{-0.1cm}
  Q(\bX=(x_1,\ldots,x_N) ; \bq) = \prod_{i = 1}^{N} Q_i(X_i=x_i ; \bq_i)\; , 
\label{eq:related:q-factorized}
\end{equation}
where $Q_i( \, . \,  ; \bq_i)$ is a categorical distribution with mean
parameters $\bq_i$. That is,
\begin{equation}
\label{eq:related:definition_q}
\forall l, \, Q_i(X_i = l ; \bq_i) = q_{i,l},
\end{equation}
with $\bq$ on the simplex $\mM$ such that $\forall i,l, \; 0 \leq q_{i,l} \leq 1$ and $\forall i,
\sum_l q_{i,l}=1$. 

$Q$ is then used to approximate $P(\bX \mid \bI)$ by minimizing the KL-divergence:
\begin{equation}
\KL(Q || P) = \sum_{\bm{x}} Q(\bX = \bx ; \bq) \log \frac{Q(\bX = \bx ; \bq)} {P(\bX = \bx \mid \bI)}\; .
\label{eq:related:kl}
\end{equation}
In  some  cases,  this  approximation  is the  desired  final
result. In others, one seeks a MAP assignment. To this end, a standard method is
to select the assignment that maximizes the \textit{approximate} posterior $Q(\bX ; \bq)$,
which is  equivalent to rounding  when the  $X_i$s are Bernoulli  variables.  An
alternative approach is to draw samples from $Q(\bX ; \bq)$.


When minimizing the KL-divergence of  Eq.~\ref{eq:related:kl}, $Q(\bX ; \bq)$ can be
reparameterized in terms of its {\it natural} parameters defined as follows. For
each variable $X_i$ and label $l$,  we take the natural parameter $\theta_{i,l}$
to be such that
\begin{equation}
  Q(X_i = l ; \bq_i) = q_{i,l} \propto \exp[-\theta_{i,l}] .
  \label{eq:naturalParam}
\end{equation}
As  we  will see  below,  this  typically yields simpler  notations  and implementations.

\vspace{-0.3cm}
\subsubsection{Sweep Mean-Field Inference}
\label{sec:related:seqMF}

Solving the minimization problem of Eq.~\ref{eq:related:kl} is equivalent~\cite{Bishop06} to minimizing
\begin{equation}
\mathcal{F}(\bq) = 
\underbrace{-\mathbf{E}_{Q(\bm{X} ; \bq)}[\log P (\bX \mid \bI)]}_{\mathcal{E}(\bq)} +
\underbrace{\mathbf{E}_{Q(\bm{X} ; \bq)}[\log Q(\bm{X} ; \bq)]}_{-\mH(\bq)},
\label{eq:related:mf-objective}
\end{equation}
with  respect to  $\bq$ on the simplex $\mM$. $\mF(.)$  is  sometimes   called  the
variational free energy. Its first term is the expectation of the energy under
$Q(\bX ; \bq)$, and its second term is the negative entropy, which acts as a regularizer.

One  can  minimize $\mathcal{F}(\bq)$  by  iteratively  updating each  $q_{i,l}$  in
sequence while  keeping the  others fixed. Each update involves setting $q_{i,l}$ to
\begin{equation}
q^{\star}_{i,l} \propto \exp\left[\mathbf{E}_{Q(\bX / X_i ; \bq)} \big[ \log P(\bX \mid \bI) \big] \right].
\label{eq:related:sweep-update}
\end{equation}

 This coordinate descent procedure, which we will call \texttt{SWEEP}, is guaranteed to
 converge to a local minimum of $\mF$~\cite{Bishop06}.
 However, it tends to be very slow for realistic image  sizes and impractical for many Computer
 Vision problems~\cite{Vineet14,Kraehenbuehl13}. Namely, in the case of dense random fields, it involves re-computing a large number of expectations
(one per factor adjacent to the variable)  
after each sequential update. Filter-based mean-field inference~\cite{Kraehenbuehl11} attempts to reduce
the complexity of these updates, but it effectively performs parallel updates, which we will describe
below. 

\vspace{-0.3cm}
\subsubsection{Parallel Mean-Field Inference}
\label{sec:parallel-mf}

To   obtain  reasonable   efficiency  in   practice,  Computer   Vision
practitioners often perform the  updates of Eq.~\ref{eq:related:sweep-update} in
parallel as opposed to sequentially. Not only does it avoid having to reevaluate
a large number of factors after each update, it also allows the use
of vectorized instructions and GPUs, both of which can have a dramatic impact on
the computation speed.


Unfortunately, these parallel updates invalidate the convergence guarantees and
in practice often lead to undesirable oscillations in the objective. Several approaches  to remedying this problem
have been proposed, which we review below.

\vspace{-0.3cm}
\paragraph{Damping}

A  natural   way  to  improve   convergence  is   to  replace  the   updates  of
Eq.~\ref{eq:related:sweep-update} by a damped version, expressed as
\begin{equation}
q_{i,l}^{t+1} = (1-\eta) \cdot q_{i,l}^{t} + \eta \cdot q_{i,l}^{\star} \; ,
\label{eq:related:parallel-update}
\end{equation}
where  $t$ denotes  the current  iteration, $q_{i,l}^{\star}$  is the  result of
solving  the  optimization  problem  of  Eq.~\ref{eq:related:sweep-update},  and
$\eta$ is a  heuristically chosen damping parameter. This damping is  explicitly mentioned in
papers     such    as~\cite{Campbell13,Sun13,Frostig14}.     In~\cite{Vineet14},
convergence  issues are  mentioned  and  a  damping parameter  is
provided  in the  publicly available  code.  
Similarly,  in~\cite{Fleuret08a}, the  algorithm relies  on
mean-field  optimization with  repulsive terms.   The  need for  damping is  not
explicitly discussed in the paper, but  the publicly available code also includes
a damping. 

Damping  delivers satisfactory  results  in  many cases,  but  does not  formally
guarantee convergence.   It may fail  if the  parameter $\eta$ is  not carefully
chosen, and sometimes  changed at different stages of the  optimization.  In all
the approaches that we are aware of,  this is done heuristically.  We will refer
to this type of methods as \texttt{ADHOC}.

\vspace{-0.3cm}
\paragraph{Concave potentials}

A principled way to address the convergence issue for the pairwise random fields is offered
in~\cite{Kraehenbuehl13}, and we refer to the corresponding algorithm as
\texttt{FULL-PARALLEL}. However, authors restrict their potentials $\phi_{ij}$ of
Eq.~\ref{eq:related:pairwise-mrf} to be concave, which in some 
cases is reasonable, but as we  will show in Section~\ref{sec:evaluation}, many Computer Vision models
violate this requirement. By contrast, our approach is similarly principled but without
additional constraints. In practice it works for higher-order, or, equivalently, non-pairwise 
potentials.

\subsection{Proximal Gradient Descent}
\label{sec:related:PGD}

Let $F$  be a generic objective  function of the form  $F(\bx)=f(\bx) + g(\bx)$,
where $g$  is a  regularizer, and  $\bx_t$ is  the value  of the
optimized  variable  at iteration  $t$  of  a minimization  procedure on a constraint set $\mX$.  Proximal
gradient descent, also known as composite mirror-descent \cite{Duchi10}, is 
an iterative method that relies on the update rule
\begin{equation}
\bx^{t+1} = \argmin_{\bx \in \mathcal{X}}\{ \langle \bx, \nabla f(\bx^t) \rangle + g(\bx) + \lambda
\Psi(\bx,\bx^t) \} \; \; , 
  \label{eq:proxUpdate}
\end{equation}
where   $\Psi$  is   a  non-negative   \textit{proximal}  function   that  satisfies
$\Psi(\bx,\bx^t) =  0$ if and only  if $\bx =  \bx^t$, and $\lambda$ is  a scalar
parameter. $g$ contains the terms of the objective function that do not need to
be approximated to the first order,
while still allowing efficient computation of
update of Eq.~\ref{eq:proxUpdate}.
  $\Psi$  can be understood as  a distance function that  accounts for
the geometry of $\mX$~\cite{Teboulle92} while also  making it possible to compute the
update of Eq.~\ref{eq:proxUpdate} efficiently. $\lambda$  can then be thought of
as the inverse  of the step size. \comment{Intuitively, solving  the minimization problem
of Eq.~\ref{eq:proxUpdate}  amounts to minimizing the  first-order approximation
of  $F(\bx)$, while  taking  into account  the composite  structure of  the
function and keeping the next  estimate $\bx^{t+1}$ close  to $\bx^t$.   }

As  shown  in  Section~\ref{sec:PGD}, our  algorithm  is  a version  of
proximal gradient  descent in  which $\Psi$  is based  on the  KL-divergence and
allows  automated step-size  adaptation as  the optimization  progresses. \comment{It  is
related to the AdaGrad method~\cite{Duchi11}, which is a gradient descent technique that uses
adaptive proximal terms, and has been extensively used for Deep Learning purposes.
However, since  AdaGrad uses a Euclidean norm as its proximal function, is not well suited for
probability distributions. } Recently, a variational approach  that also relies on  the KL-divergence
as  the proximal function has been  proposed~\cite{Emti15}. This paper explores  the
connection between the KL-proximal  method and the Stochastic Variational
Inference~\cite{Amari98,Hoffman13}.  However,  it does  not allow for  step size
adaptation, which often yields better performance, as we demonstrate in our
experiments and is not directly applied to the mean-field setting.

%% file: method.tex

As discussed in the previous section, the goal of mean-field inference is to
\begin{equation}
\label{eq:method:minimize-mf}
\underset{\bq \in \mM}{\text{minimize }} \mF(\bq)
\end{equation}
where $\mF$ is the variational free energy of Eq.~\ref{eq:related:mf-objective}.
Performing sequential updates of the $q_{i,l}$ is guaranteed to converge, but can
be slow. Parallel updates are usually much faster, but the optimization procedure
may fail  to converge.  \comment{ This can  often be prevented by  introducing a carefully
weighted damping  term but without formal  proof of convergence.  Only  when the
pairwise  potentials  of  Eq.~\ref{eq:related:pairwise-mrf} are  concave  has  a
principled way to achieve convergence been proposed~\cite{Kraehenbuehl13}, and it
only applies to that specific situation.}

In this section, we introduce  our approach to guaranteeing convergence whatever
the shape of the pairwise potentials. To  this end, we rely on proximal gradient
descent  as  described  in  Section~\ref{sec:PGD}  and  formulate  the  proximal
function $\Psi$  in terms of  the KL-divergence. This  is motivated by  the fact
that  it  is  more  adapted   to  measuring  the  distance  between  probability
distributions  than  the usual  L2  norm,  while  being  independent of  how  the
distribution is parameterized.

We will show  that this both guarantees convergence and  yields a principled way
to obtain a closed form damped update equation equivalent to Eq.~\ref{eq:related:parallel-update}.

\subsection{Proximal Gradient for Mean-Field Inference}
\label{sec:PGD}

In   our   approach    to   minimizing   the   variational    free   energy   of
Eq.~\ref{eq:related:mf-objective}, we treat $\mathcal{E}$ as the function $f$ of
Eq.~\ref{eq:proxUpdate}  and  the negative  entropy  $-\mH$  as the  regularizer
$g$.  This choice  stems from  the  fact that  $-\mH$ is  separable,  and
therefore, can be minimized in parallel in 
Eq.~\ref{eq:proxUpdate}, without using a first order approximation.
Also, $-\mH$ being  the regularizer  $g$ means
that  we  do  not  need  to look at  its  derivatives  with  respect  to  the
mean-parameters, which are not well behaved when they approach zero. We then define

\begin{equation}
  \Psi^t(\bq,\bq^t) = \sum  \limits_i \sum \limits_l d^t_{i,l} q_{i,l} \log \dfrac{q_{i,l}}{q_{i,l}^t}
 = \bD^t \odot \KL(\bq || \bq^t) \; , \label{eq:psiQ}
\end{equation}
%
where $\KL$ is  the non-negative KL-divergence, which is a  natural choice for a
distance  between distributions.   $\bD^t$ is  a diagonal  matrix with  positive
diagonal elements $d^t_{i,l}$s, which we  introduce to allow for anisotropic scaling
of the proximal KL-divergence term. As will be discussed below,  different choices of
the $d^t_{i,l}$s  yield different  variants of our  algorithms.  Note  however that,
$\Psi^t$ is a valid proximal function.
\comment{whatever their values, $\Psi^t$ is a valid proximal function as long as they are
all non-negative since $\Psi(\bq,\bq^t)=0$ if and only if $\bq = \bq^t$.}

The update of Eq.~\ref{eq:proxUpdate} then becomes
\begin{equation}
\label{eq:method:proxUpdateOurs}
\bq^{t+1} = \argmin_{\bq \in \mM} \{
\langle  \bq,  \nabla  \mE(\bq^t)  \rangle  - \mH(\bq)  +  \bD^t  \odot  \KL(\bq
||\bq^{t}) \}  \; .
\end{equation}
This  computation  can  be  performed   independently  for  each  index  $i  \in
\{1,\dots,N\}$. Furthermore, we prove in  the supplementary material that it can
be done in closed form and can be written as
\begin{eqnarray}
q_{i,l}^{t+1} & \propto \exp[ & \eta^t_{i,l} \cdot \mathbf{E}_{Q(\bX  /  X_i=l ; \bq)}  \big[ \log  P(\bX  | \bI)  \big] 
 \label{eq:method:update_mean}  \\
   & & + (1-\eta^t_{i,l}) \cdot \log q_{i,l}^t ] \; , \nonumber
\end{eqnarray}
where  $\eta^t_{i,l} =  \dfrac{1}{1+d^t_{i,l}}$.  Eq~\ref{eq:method:update_mean}  can be
rewritten as
\begin{equation}
\label{eq:method:update_natural} 
\theta_{i,l}^{t+1} = \eta^t_{i,l} \cdot \theta_{i,l}^{\star} + (1-\eta^t_{i,l}) \cdot \theta_{i,l}^{t} \; \; ,  
\end{equation}
where $\theta_{i,l}^{\star} = -\mathbf{E}_{Q(\bX /  X_i ; \bq)} \big[ \log P(\bX | \bI)
  \big]$ now is a natural parameter, like those of Eq.~\ref{eq:naturalParam}. In
other   words,    we   have    replaced   the    heuristic   update    rule   of
Eq.~\ref{eq:related:parallel-update}  in  the  space  of mean  parameters  by  a
principled  one  in the  space  of  natural ones.  As  we  will see,  this
yields performance and convergence improvements in most cases. 
As for the stopping criteria, one can define one based on the value of the objective, or, in
practice, run inference for a fixed number of iterations.

\subsection{Fixed Step Size}
\label{sec:fixedStep}

The simplest way to  instantiate our algorithm is to fix  all the $d^t_{i,l}$s of
Eq.~\ref{eq:psiQ} to the same value $d$ and to write
\begin{equation}
  \forall t  \; ,  \; \bD^t =  \bD =  d \mathbb{I} \;  \Rightarrow \;  \forall t,i,l,
  \eta^t_{i,l} = \dfrac{1}{1+d} \; ,
  \label{eq:fixedStepSize}
\end{equation}
where  $\eta^t_{i,l}$   plays  the  same   role  as   the  damping  factor of
  Eq.~\ref{eq:related:parallel-update}. \comment{In fact, the whole update rule is similar
    to the one of Eq.~\ref{eq:related:parallel-update}, except for the fact that
    the damping is performed in the  space of natural parameters instead of mean
    ones. } We  now show that  this is guaranteed  to converge when  the proximal
    term is given enough weight.

In  our mean-field  settings,  $\mE(\bq)$   is  a  polynomial  function   of  the
mean-parameters vector $\bq$.  Therefore, one can always find some positive real
number $L$ such that $\mE$ is {\it L-Lipschitz gradient}, which means
\begin{equation}
 \forall x,y \in \mM \; , \; \|\nabla \mE(x) - \nabla \mE(y)\|_2 \leq L \|x-y \|_2 \; .
\end{equation}
In the  supplementary material,  we prove  that this  property implies  that our
proximal gradient descent scheme is guaranteed  to converge for any fixed matrix
$D = d \mathbb{I}$ such that $d  >L $.
\comment{

  \begin{theorem}[Convergence]
Let us assume that the $\mE(q)$ is L-Lipschitz gradient for some real constant $L$. 
Our 
  \end{theorem} 
  
  \begin{proof}
  The proof is provided in the supplementary material.
  \end{proof}
}

Intuitively, when updating the value  of $\bq^{t}$ to $\bq^{t+1}$, the magnitude
of  the   gradient  change   controlled  and thus  the   coordinate-wise  optimum
$\theta_{i,l}^{\star} = - \nabla \mE(\bq^t)_{i,l}$ will also be changing smoothly
across  iterations. As a  result,  $L$   is  the  key  value to understand
  oscillations.  In practice, our goal is to find its smallest possible value to
  allow steps as large as possible while guaranteeing convergence.

In the pairwise case, the Hessian of the objective function is a constant matrix, which
we call potential matrix. Therefore, the highest eigenvalue  of the potential matrix  is a valid
Lipschitz constant and efficient methods  allow to compute it for moderately sized problems.

In fact, the  convergence result presented in~\cite{Kraehenbuehl13} is strongly related to this.
Namely, assuming that the potential matrix is negative semi-definite, is equivalent to assuming that
$L<0$ in our formulation. This directly corresponds to the concavity assumptions on the potentials
in~\cite{Kraehenbuehl13}. Therefore, under  the assumptions   of~\cite{Kraehenbuehl13},
our algorithm leads to $\eta=1$, corresponding  to  the fully-parallel  update  procedure.
In that sense, our procedure is a generalization of the one proposed by~\cite{Kraehenbuehl13}. 

In the non-pairwise case, the Hessian is not constant, and the calculation of the Lipschitz constant
is not trivial. For each specific problem, bounds should be derived using the particular
shape of the CRF at hand.

\subsection{Adaptive Step Size}
\label{sec:method:qs}

Note that the Hessian of the KL-proximal term is diagonal with
\begin{equation}
\label{eq:method:hessian0} 
\dfrac{\partial^2 \bD^t \cdot 
\KL(\bq ||\bq^{t})}{\partial q_{i,l}^2}|_{\bq=\bq^{t}}    = \dfrac{d^t_{i,l}}{q^{t}_{i,l}}\; .
\end{equation}
Therefore, when  some of  the $q_{i,l}$s  get close  to 0,  the elements  of the
Hessian may become very  large, especially when using a  constant value for the
$d^t_{i,l}$ as  suggested above.   When that  happens, the  local KL-approximation
remains a valid  upper bound of the  objective function, but not  a tight enough
one,  which results  in step  sizes  that are  too small  for fast  convergence.

This can  be reduced  by choosing  a matrix  $\bD^t$ that  compensates for
this.    A    simple   way   to   do    this   would   be   to   scale the $d_{i,l}^t $ proportionally to 
$\max(q_{i,0},\dots,q_{i,L_i-1})$  to start compensating for diagonal terms.  However,  this  method is  still  sub-optimal
because  it ignores  the fact  that  all our  variables lie  inside the  simplex
$\mM$. A better
alternative is to bound from below the proximal term by a quadratic function, but on $\mM$
rather than on $ \mathbb{R}^n$. 

In this paper, we only apply this method to the binary case, for which we set
\begin{equation}
\label{eq:method:hessian} 
d^t_{i,0} =d^t_{i,1}= q^t_{i,0}q^t_{i,1} \cdot d \; ,
\end{equation}
were $d$ is an additional parameter that should be set close to $L$. Extending this approach
to the multi-label case will be a topic for future work. In Section~\ref{sec:momentum}, we provide a
different alternative to performing adaptive anisotropic updates in all settings. 

Intuitively, when the current parameters are close to the
borders of the simplex, the mean parameters are less sensititive to natural parameters, which, therefore,
need less damping.
We demonstrate in our experiments that it provides a way to choose the step size without 
tuning. 

\subsection{Momentum}
\label{sec:momentum}

Our approach can easily be extended  to incorporate techniques that are known to
speed-up  gradient descent  and help to avoid local  minima, such  as the  classic
momentum     method~\cite{Polyak64}     or      the     more     recent     ADAM
technique~\cite{Kingma14}.
The  momentum  method  involves  averaging the  gradients  of  the  objective
 $f(\bx)$ over the iterations in a \textit{momentum} vector $\bmm$ and use it as the direction for the update instead of simply following the current gradient. 
To  integrate it into our  framework, we replace
the gradient  $\nabla \mE$ in Eq.~\ref{eq:method:proxUpdateOurs}  by its rolling exponentially weighted
average $\bmm$ computed as
\begin{eqnarray}
\bmm^{t+1} & = & \gamma_1 \bmm^t + (1 - \gamma_1) \nabla \mE(\bq^t) \; , 
\end{eqnarray}
with the exponential decay parameter $\gamma_1 \in [0;1]$. This substitution brings the following update rule
\begin{eqnarray}
\label{eq:momentumUpdate}
\theta_{i,l}^{t+1} & = & \eta \cdot m_{i,l}^t+ (1-\eta) \cdot \theta_{i,l}^{t}  \; .
\end{eqnarray}
We will refer to this approach as \texttt{OURS-MOMENTUM}.

\subsection{ADAM}
\label{sec:adam}

The ADAM method~\cite{Kingma14} has become very popular in deep learning. Our
framework  makes  it   easy  to  use  for  mean-field inference   as  well  by
appropriately choosing the matrix $\bD^t$ at each step and combining it with the momentum technique. 

We define the averaged second moment vector $\bv$ of the natural gradient as
\begin{eqnarray}
{v}^{t+1}_{i,l} & = &\gamma_2 [\theta^t_{i,l} +\nabla \mE(\bq^t)_{i,l}]^2 + (1 - \gamma_2) {v}^t_{i,l} \; , 
\end{eqnarray}
where $\bv$ is initialised to a strictly positive value and $\gamma_2 \in [0;1]$ is an exponential
memory parameter for $\bv$. 

Then, the $\bD^t$ matrix is defined through each of its diagonal entries as
\begin{equation}
  d^t_{i,l} = \sqrt{{v}^{t+1}_{i,l}}d +\epsilon - 1 \;,
  \label{eq:adamStepSize}
\end{equation}
where $\epsilon$ is a fixed parameters and $d$ controls the damping. We will refer to this method as \texttt{OURS-ADAM}.  

Intuitively it is good at exploring parameter space thanks to a form of auto-annealing
of the gradient. The natural gradient $\btheta_t + \nabla \mE(\bq^t)$ is zero at a local minimum of
the objective function~\cite{Hoffman13}. Therefore, close to a minimum, the proximal term $\bD^t$
becomes small, thus allowing more exploration of the space. On the other hand, after a long period
of exploration with large natural gradients, more damping will tend to make the algorithm converge.

%% file: evaluation.tex

In this  section, we evaluate our method on a variety  of inference  problems  and
demonstrate  that  in  most  cases  it  yields  faster convergence and better  minima.  
All the code, including our efficient GPU mean-field inference framework, will be made publicly
available. 

\subsection{Baselines and Variants}

We  compare  several variants  of  our  approach to  some  of  the baselines  we
introduced  in the  related  work  section. The  baselines  we  consider are  as
follows:
 \begin{itemize}
  \setlength{\itemsep}{1pt}
  \setlength{\parskip}{0pt}
  \setlength{\parsep}{0pt}
   \item  \texttt{SWEEP}. As  discussed  in Section~\ref{sec:related:seqMF},  it
     involves  sequential  coordinate  descent~\cite{Bishop06} and  is  not
     always computationally tractable for large problems.
   \item  \texttt{ADHOC}.   As  discussed in  Section~\ref{sec:parallel-mf},  it
     performs parallel updates with the {\it ad hoc} damping parameter $\eta$ of
     Eq.~\ref{eq:related:parallel-update} chosen manually. 
   \item      \texttt{FULL-PARALLEL}.        As      also       discussed      in
     Section~\ref{sec:parallel-mf}, it  relies on the inference described in~\cite{Kraehenbuehl13}. 
     For example, the popular \texttt{densecrf} framework~\cite{Kraehenbuehl11} uses this approach.  
\end{itemize}
We compare to these the following variants of our approach:
\vspace{-0.3cm}
\begin{itemize}
  \setlength{\itemsep}{1pt}
  \setlength{\parskip}{0pt}
  \setlength{\parsep}{0pt}

  \item \texttt{OURS-FIXED}. Damping  occurs in the space  of natural parameters
    instead  of mean  ones as  described in  Section~\ref{sec:fixedStep}. 
  \item \texttt{OURS-ADAPTIVE}. Adaptative  and anisotropic damping in the
    space       of      natural       parameters      as       described      in
    Section~\ref{sec:method:qs}. 
  \item \texttt{OURS-MOMENTUM}. Similar to  \texttt{OURS-ADAPTIVE}, but using the
    momentum  method  instead of  ordinary  gradient  descent, as  described  in
    Section~\ref{sec:momentum}.  We use  the  same parameter  value $\gamma_1  =
    0.95$ for all datasets.
  \item \texttt{OURS-ADAM}. Similar to \texttt{OURS-ADAPTIVE} but using the ADAM
    method    instead of ordinary gradient as  described   in
    Section~\ref{sec:adam}. We  use the same  parameters as in  the original
    publication~\cite{Kingma14}, $\gamma_1  = 0.99$, $\gamma_2 =  0.999$ and $\epsilon =$1E-8 for
    all datasets. 
\end{itemize}
All   four   methods    involve   a  parameter    $\eta = \frac{1}{1 + d}$,   defined   in
Eq.~\ref{eq:fixedStepSize} for \texttt{OURS-FIXED}, Eq.~\ref{eq:method:hessian} for
\texttt{OURS-ADAPTIVE}, Eq.~\ref{eq:momentumUpdate}   for   \texttt{OURS-MOMENTUM}
and Eq.~\ref{eq:adamStepSize} for \texttt{OURS-ADAM}. Additionally, in
Section~\ref{sec:eval:results} and Fig.~\ref{fi:eval:benchmarks:steps} we demonstrate that our
method is less sensitive to the choice of this parameter than its competitors. 


\subsection{Experimental Setup}

We   evaluated  all   these    methods   first   on    a   set    of   standardized
benchmarks~\cite{Frostig14}: \texttt{DBN}, 
containing 108 instances of deep belief networks, \texttt{GRID},
containing 21 instances of two-dimensional grids, and \texttt{SEG},
containing 100 instances of segmentation problems, where each instance is
represented as a binary pairwise random field. 


We then consider three realistic Computer Vision tasks that all involve minimizing a
functional of the form given in Eq.~\ref{eq:related:mf-objective}. We describe them below.


\vspace{-0.35cm}
\paragraph{Characters  Inpainting} We consider character inpainting, formulated as a
binary pairwise random field, Decision Tree Fields (DTF,~\cite{Nowozin11}). 
The dataset contains 100 test instances of occluded characters, and the goal is to restore the
occluded part, as shown in the last row of Fig.~\ref{fig:intro:teaser}.   
We use pre-computed potentials provided by the authors of~\cite{Nowozin11}. Note, that this model
consists of data-driven potentials, and includes both short and long-range interactions, which
makes it particularly interesting from the optimization perspective.  

\vspace{-0.35cm}
\paragraph{People  Detection} 
We consider detecting upright people in a multi-camera settings, using the Probabilistic Occupancy Map
approach (POM,~\cite{Fleuret08a}), that relies on a random field with high-order repulsive
potentials, which models background subtraction signal given the presences of people in the
environment.  
We evaluate it on the ISSIA~\cite{DOrazio09} dataset, which contains 3000 frames of a
football game, captured by 6 cameras located on two sides of the field. The original
work~\cite{Fleuret08a} does not explicitly mention it, but the publicly available  
implementation uses the~\texttt{ADHOC} damping method. 
We implement all our methods and remaining baselines directly in this code of~\cite{Fleuret08a}. \comment{In
order to make the  it tractable, we had to introduce an additional data structure, denoting this modified
algorithm \texttt{SWEEP}. }

\vspace{-0.35cm}
\paragraph{Semantic Segmentation} 
We consider semantic segmentation on PASCAL VOC 2012 dataset~\cite{Pascal-voc-2012},
which defines 20 object classes and 1 backround class. We based our evaluation on DeepLab-CRF
model~\cite{Chen15b}, which is currently one of the best-performing methods. This
model uses CNNs to obtain unary potentials, and then employs \texttt{densecrf}
of~\cite{Kraehenbuehl13} with dense pairwise potentials. However, this basic CRF model does not  
contain any strong repulsive terms, and thus we expect \texttt{densecrf}'s standard inference,
\texttt{FULL-PARALLEL}, to work well. To improve performance, we additionally introduced
co-occurrence potentials~\cite{Vineet14}, which, as we will show, violate the conditions assumed in
\texttt{densecrf}, but can still be successfully handled by our method. Intuitively, these
co-occurrence terms put priors on the sets of classes that can appear together. We made minor
modifications of \texttt{densecrf} to support both our inference and co-occurrence potentials.


%

\subsection{Comparative Results}
\label{sec:eval:results}

In order to understand how the methods behave in practical settings, when the available
computational time is limited, we evaluate all methods for several computational \textit{budgets}. 
The shortest budget corresponds to the early-stopping scenario after few iterations, the longest one
roughly models the time until convergence, and the middle one is around 20-30\% of the longest.

\vspace{-0.3cm}
\paragraph{Benchmarks}

\begin{table*}[hbt!]
\small
\begin{center}
\begin{tabular}{l?ccc?ccc?ccc}
& \multicolumn{3}{c?}{\sttt{DBN}} 
& \multicolumn{3}{c?}{\sttt{GRID}} 
& \multicolumn{3}{c}{\sttt{SEG}} \\\hline
method
&  0.05s &  0.30s &  1.00s 
&  0.05s &  0.30s &  1.00s 
&  0.05s &  0.30s &  1.00s \\
\hline
\sttt{SWEEP}         
&    -112.94 &   -2088.07 &   -2138.13 
&   -5540.59 &  -16675.55 &  -18592.26
&      78.81 &      75.50 &      75.50 
\\                       
\sttt{FULL-PARALLEL} 
&   -1952.52 &   -1951.54 &   -1942.86 
&   -2564.39 &   -2777.33 &   -2439.08
&      75.66 &      75.66 &      75.66 
\\
\sttt{ADHOC}         
&   -2047.31 &   -2047.31 &   -2047.31 
& -18345.42 &  -18348.80 &  -18349.03
&     76.10 &      75.66 &      75.66 
\\                           
\hline
\sttt{OURS-FIXED}        
&   -2081.91 &   -2081.91 &   -2081.91 
&  -18213.81 &  -18219.42 &  -18219.45 
&      77.17 &      75.61 &      75.61 
\\
\sttt{OURS-ADAPTIVE}            
&   -2125.48 &   -2130.61 &   -2130.61 
& -18245.93 &  -18252.48 &  -18252.48
& 77.68 &      75.64 &      75.61
\\
\sttt{OURS-MOMENTUM}      
&   \textbf{-2260.98} &   \textbf{-2362.14} &   \textbf{-2374.51}
& -18143.48 &  \textbf{-19074.45} &  \textbf{-19184.37}
&      74.35 &      73.75 &      73.75 
\\
\sttt{OURS-ADAM}          
&   -2107.98 &   -2107.93 &   -2107.93 
&  \textbf{-18617.06} &  -18732.59 &  -18740.36
&      \textbf{72.37} &      \textbf{72.32} &      \textbf{72.32} \\
\hline
\end{tabular}
\caption{Results for KL minimization for three benchmark datasets~\cite{Frostig14}: \texttt{DBN} (deep belief
  networks), \texttt{GRID} (two-dimensional grids), \texttt{SEG} (binary segmentation). All the
  numbers are KL divergence (lower is better) averaged over the instances.} 
\vspace{-0.5cm}
\label{tab:eval:benchmark}
\end{center}
\end{table*}

\begin{figure*}[ht!]
\begin{center}
\begin{tabular}{c@{}c@{}c@{}c@{}c@{}c}
\setlength{\tabcolsep}{0pt}
\includegraphics[trim=10mm 10mm 10mm 10mm, clip, width=0.16\textwidth]{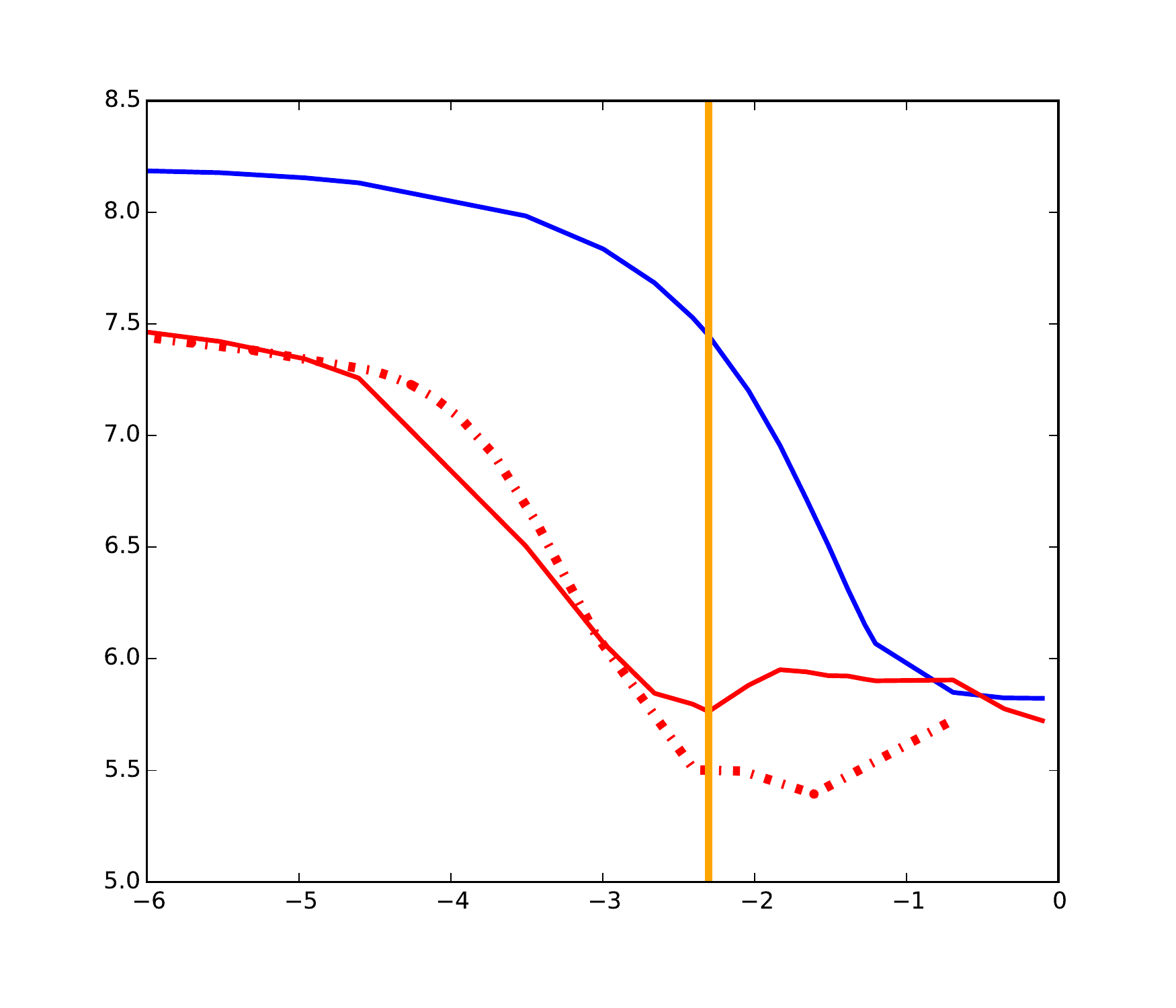} & 
\includegraphics[trim=10mm 10mm 10mm 10mm, clip, width=0.16\textwidth]{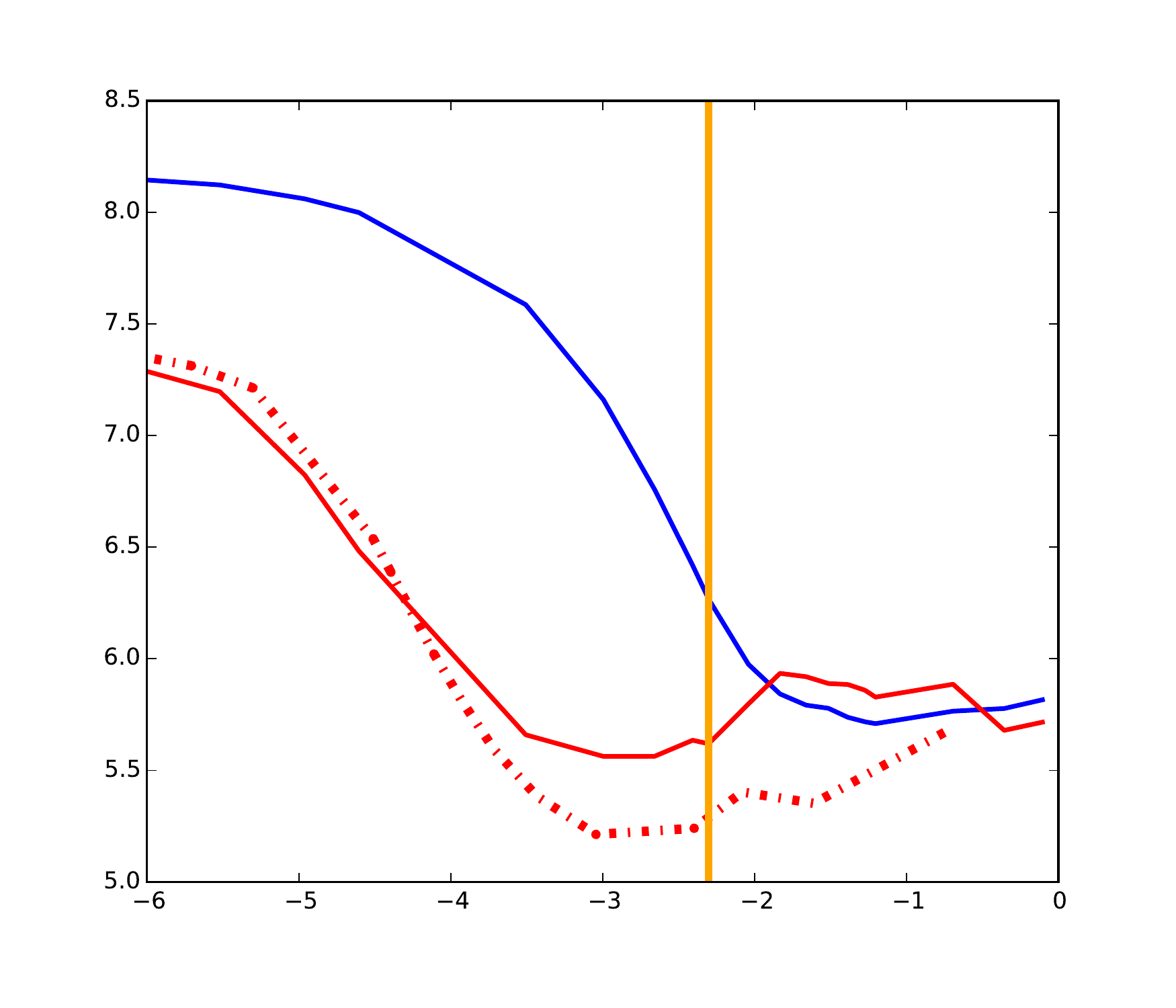} &
\includegraphics[trim=10mm 10mm 10mm 10mm, clip, width=0.16\textwidth]{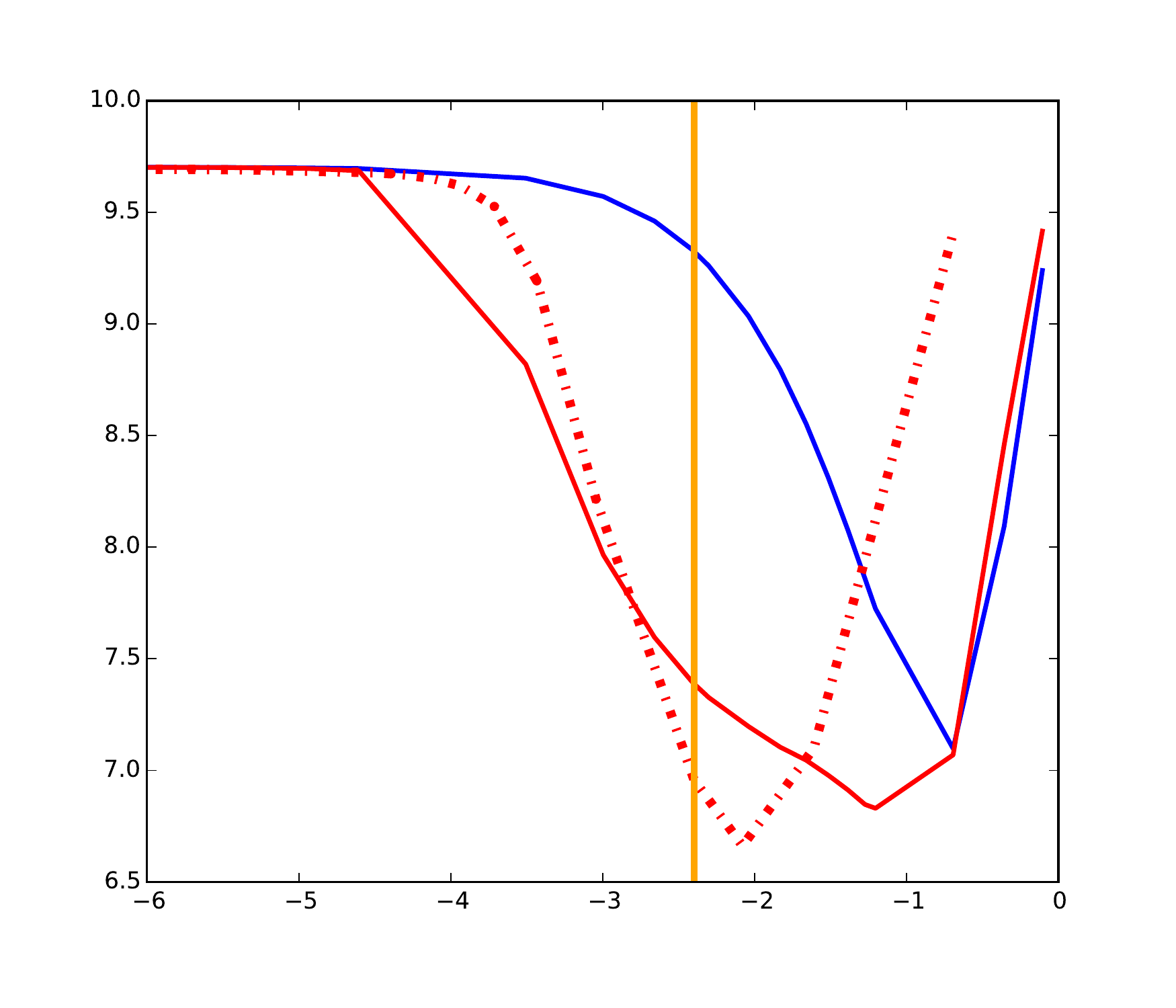} &
\includegraphics[trim=10mm 10mm 10mm 10mm, clip, width=0.16\textwidth]{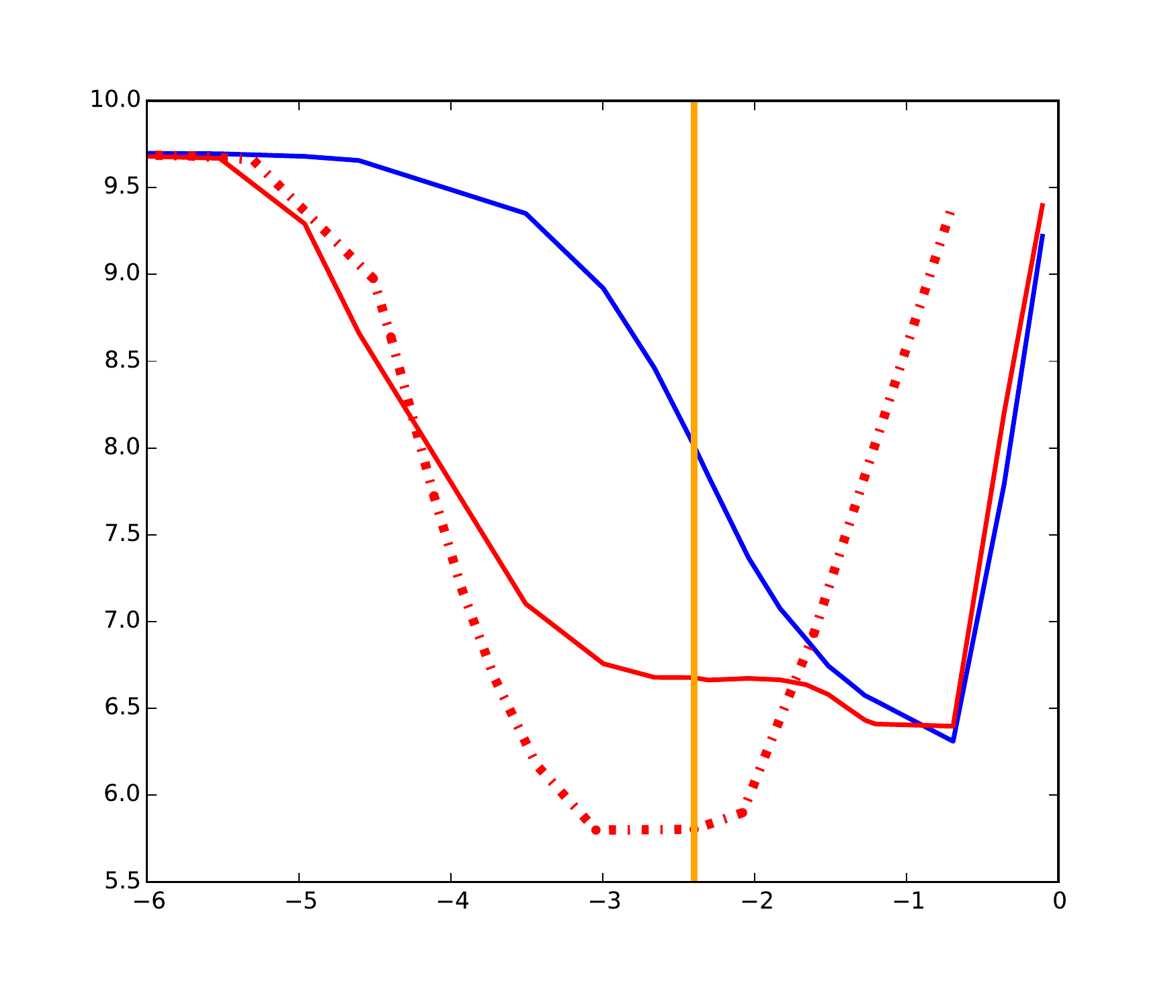} &
\includegraphics[trim=10mm 10mm 10mm 10mm, clip, width=0.16\textwidth]{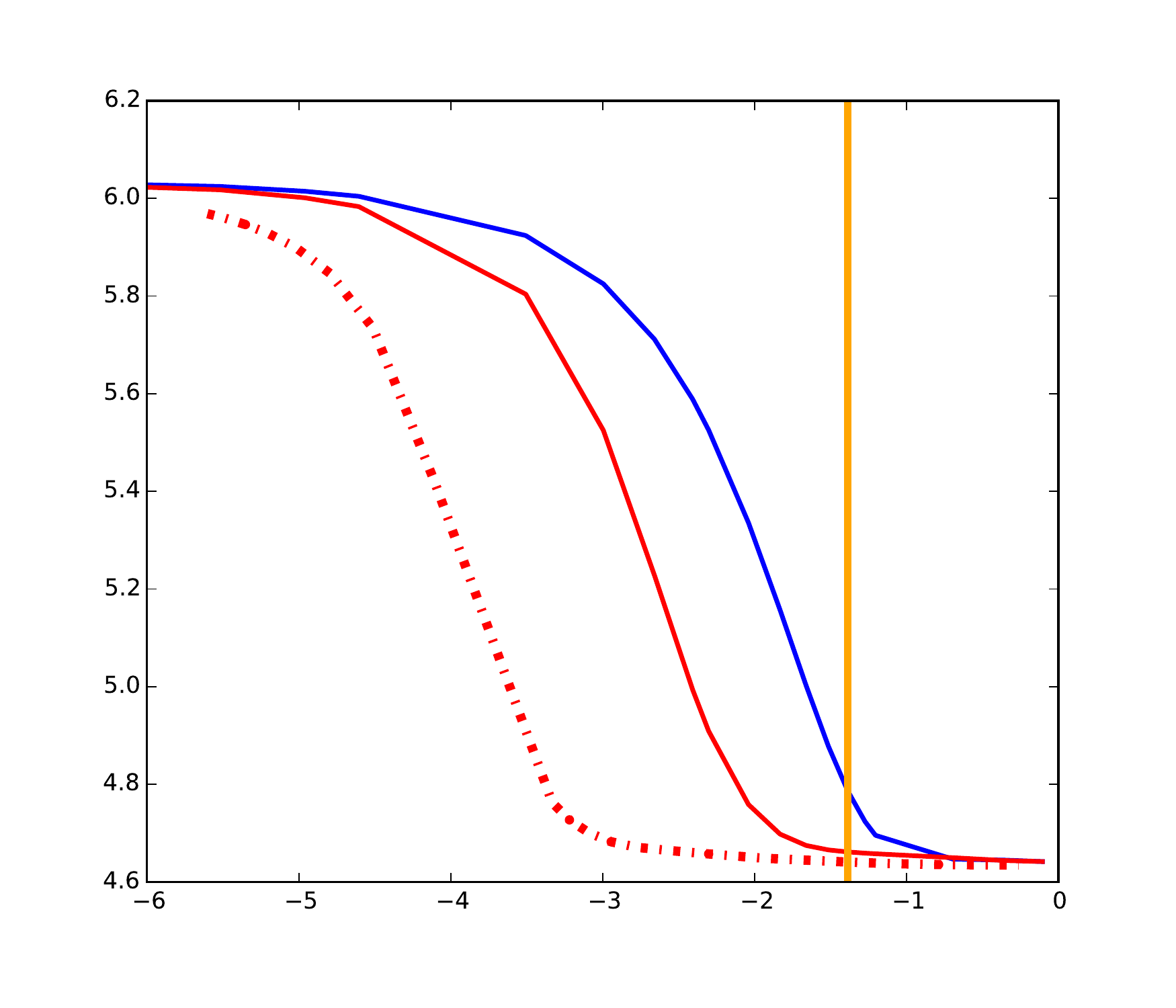} &
\includegraphics[trim=10mm 10mm 10mm 10mm, clip, width=0.16\textwidth]{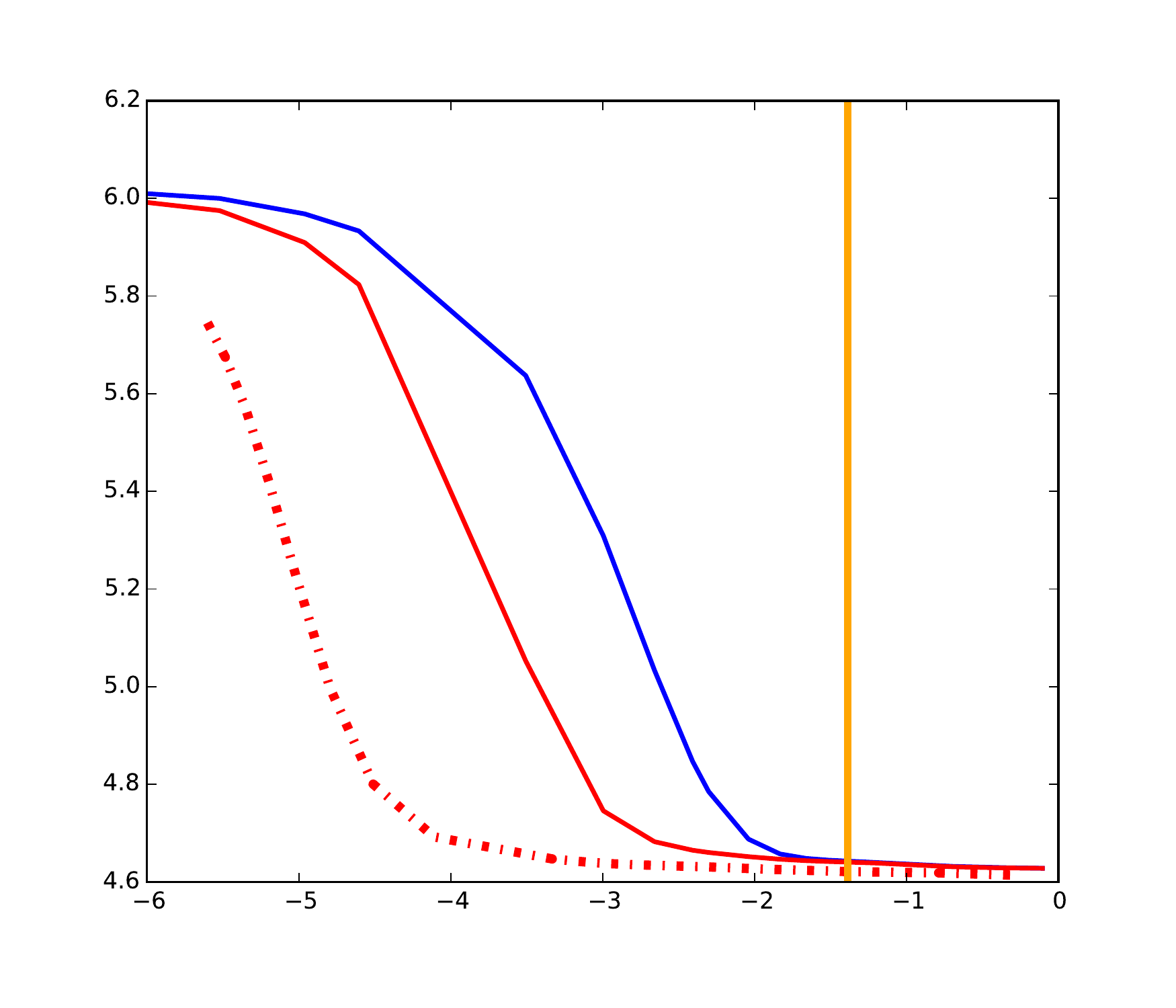} \\ 
  \sttt{ DBN(0.05s)} & \sttt{ DBN(0.3s)} &
  \sttt{ GRID(0.05s)} & \sttt{ GRID(0.3s)} &
  \sttt{ SEG(0.05s)} & \sttt{ SEG(0.3s)} \\
\end{tabular}
\end{center}
\vspace{-0.2cm}
   \caption{Insensitivity of~\texttt{OURS-FIXED} (red) and~\texttt{OURS-ADAPTIVE} (dashed red) vs
     \texttt{ADHOC} (blue) to the damping parameter \small{$\eta = \frac{1}{1+d}$}. We report
     KL-divergence (lower is better) vs the value of the parameter, both in log-space.} 
\label{fi:eval:benchmarks:steps}
\vspace{-0.3cm}
\end{figure*}

\begin{figure*}[ht!]
  \begin{center}
    \begin{tabular}{c@{}c@{}c}
 \includegraphics[trim=0mm 2mm 11mm 10mm, clip,
      width=0.3\textwidth]{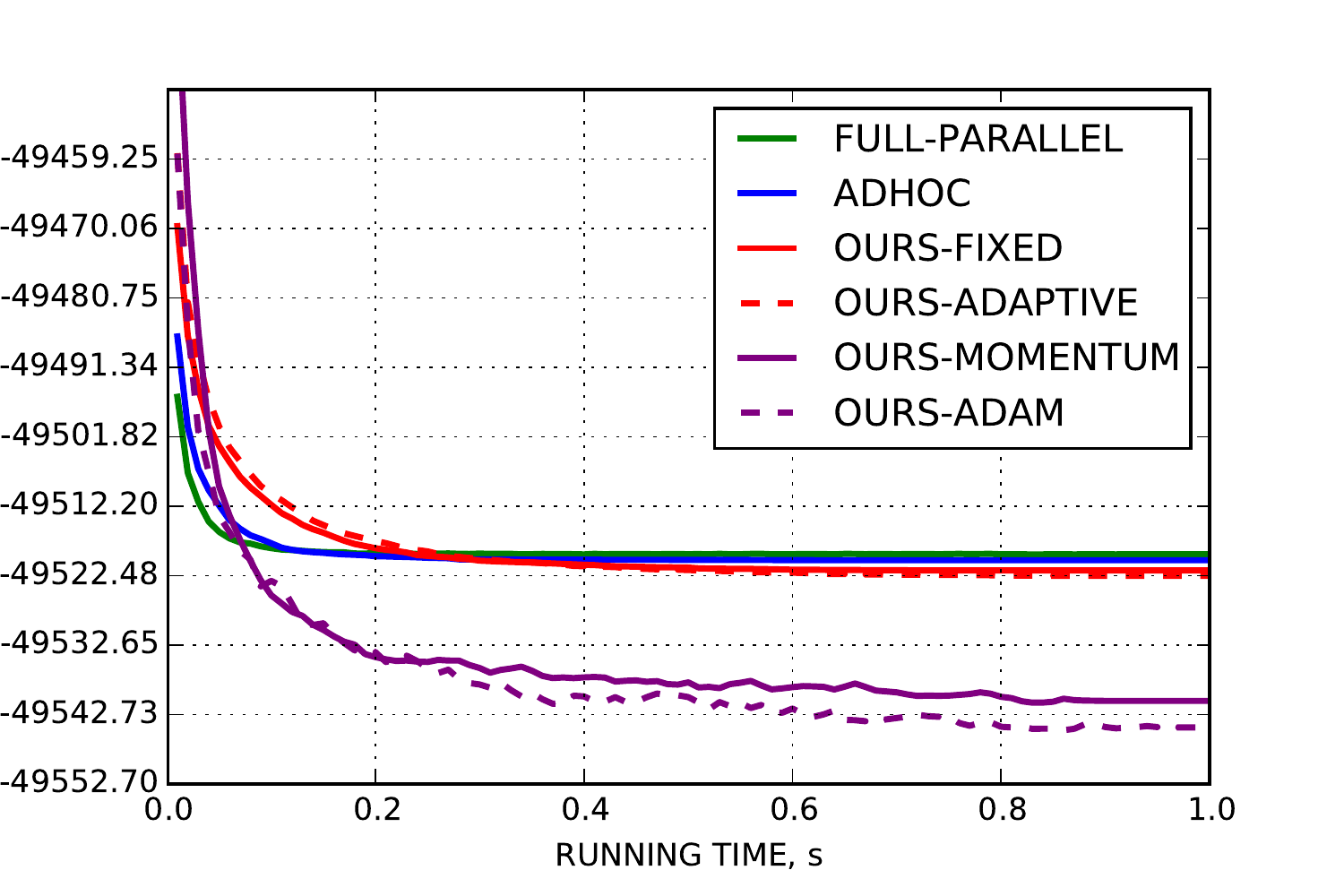}  & 
 \includegraphics[trim=2mm 2mm 11mm 10mm, clip, width=0.3\textwidth]{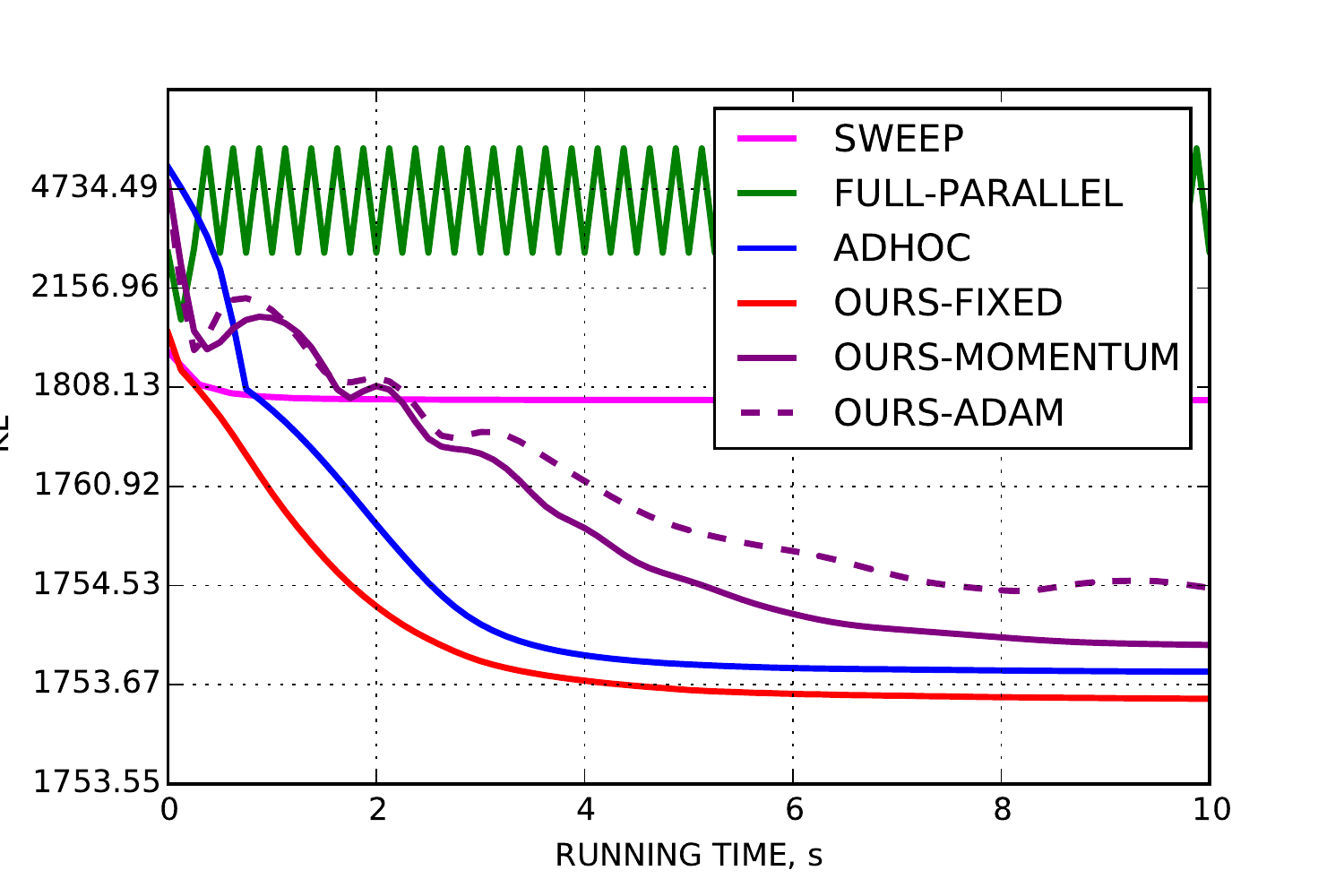}
      & 
 \includegraphics[trim=0mm 2mm 11mm 10mm, clip, width=0.3\textwidth]{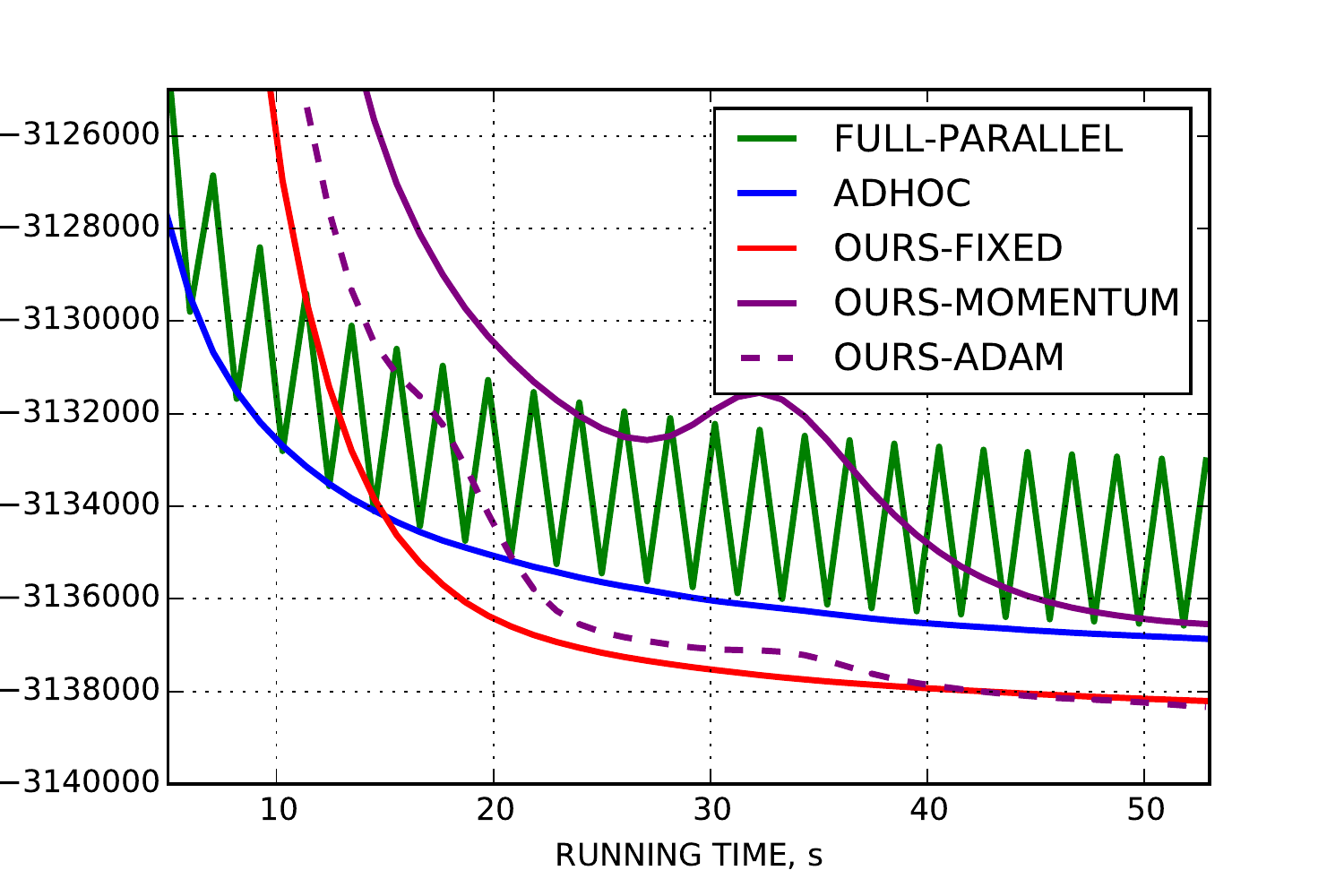}
      \\              
 (a) Characters Inpainting  &
 (b) People Detection &
 (c) Semantic Segmentation \\             

    \end{tabular}
  \end{center}
  \vspace{-0.2cm}
  \caption{Convergence results. (a) \texttt{OURS-ADAM} and \texttt{OURS-MOMENTUM} converge very fast
    to a much better minima. 
    (b) \texttt{OURS-FIXED} outperforms \texttt{ADHOC} both in terms of speed of convergence and the
    value of the objective.
    (c) \texttt{OURS-ADAM} and \texttt{OURS-FIXED} show the best performance. The former converges a bit
    slower, but in the end provide slightly better minima. \texttt{ADHOC} for this dataset converges
    rather fast, but fails to find a better optima.}  
\vspace{-0.3cm}
 \label{fi:eval:convergence}
 \end{figure*}

\begin{table*}[!hbt]
\small
\vspace{-0.25cm}
\begin{center}
\begin{tabular}{l?C{1.3cm}C{0.9cm}?C{1.3cm}C{0.9cm}?C{1.3cm}C{0.9cm}}
 & \multicolumn{2}{c?}{0.05s}  & \multicolumn{2}{c?}{0.3s} & \multicolumn{2}{c}{3s} \\\hline
method & \sttt{KL} & \sttt{PA} & \sttt{KL} & \sttt{PA} & \sttt{KL} & \sttt{PA}\\\hline
\sttt{SWEEP} & -6342.56  & 54.57  & -25233.54 & 58.38 & -49519.33 & 62.50 \\
\sttt{FULL-PARALLEL} & \textbf{-49516.98} & 60.99  & -49519.27 & 62.00 & -49519.33 & 62.05 \\
\sttt{ADHOC} & -49514.27 & 61.46  & -49520.09 & 62.15 & -49520.20 & 62.17 \\\hline
\sttt{OURS-FIXED} & -49505.59 &     60.99 & -49520.33 &    62.26 & -49521.71  &    62.35 \\
\sttt{OURS-ADAPTIVE} & -49503.43&    60.93 & -49520.14 &    62.32 & -49522.49  &    62.60 \\
\sttt{OURS-MOMENTUM} & -49513.57 &    63.69 & -49536.67 &    65.26 & -49540.76   &    65.95 \\
\sttt{OURS-ADAM} & -49516.02 & \textbf{65.36}  & \textbf{-49538.84} & \textbf{67.03} & \textbf{-49544.58} & \textbf{67.12} \\\hline
\end{tabular}
\end{center}
\vspace{-0.2cm}
\caption{Results for characters inpainting
  problem~\cite{Nowozin11} based on DTFs. \texttt{PA} is the pixel accuracy for the occluded
  region (bigger is better). Our methods outperform the baselines by a margin of 3-5\%. Since
  \texttt{FULL-PARALLEL} is not damped, it gets to low KL-divergence value quickly, however the
  actual solution is significantly worse. }  
\label{tab:eval:dtf}

\vspace{-0.25cm}
\begin{center}
\begin{tabular}{l?C{1.3cm}C{0.9cm}?C{1.3cm}C{0.9cm}?C{1.3cm}C{0.9cm}}
 & \multicolumn{2}{c?}{0.5s}  & \multicolumn{2}{c?}{1.3s} & \multicolumn{2}{c}{5s} \\\hline
method & \sttt{KL}&\sttt{MODA}&\sttt{KL}&\sttt{MODA}& \sttt{KL}& \sttt{MODA}\\ 
\hline
\sttt{SWEEP}   & 1865.43 & \textbf{0.630}  & 1795.66 & 0.656 & 1795.60 & 0.656\\
\sttt{FULL-PARALLEL} & 2573.79 & 0.000  &2573.79 & 0.000 & 8500.90 & 0.030\\
\sttt{ADHOC} & 2573.79 & 0.308 & 1760.02  & 0.781 & 1753.71 & \textbf{0.829} \\\hline
\sttt{OURS-FIXED} &  \textbf{1783.63} & 0.626 & \textbf{1754.55}  & \textbf{0.802} &  \textbf{1753.63} & \textbf{0.829}\\
\sttt{OURS-MOMENTUM} & 1931.36 & 0.040  & 1797.19  & 0.650 &  1753.83 & 0.826\\
\sttt{OURS-ADAM} &  2008.52 & 0.021  & 1813.66  & 0.501 & 1754.52 & 0.824\\
\hline
\end{tabular}
\end{center}
\vspace{-0.2cm}
\caption{Results for people detection task~\cite{DOrazio09} based on
  POM~\cite{Fleuret08a}. \texttt{OURS-FIXED} outperforms the baselines and adaptive methods. This
  means that this problem does not require more sophisticated parameter exploration techniques. }  
\label{tab:eval:pom-soccer}

\vspace{-0.25cm}
\begin{center}
\begin{tabular}{l?C{1.3cm}C{0.9cm}?C{1.3cm}C{0.9cm}?C{1.3cm}C{0.9cm}}
 & \multicolumn{2}{c?}{5s}  & \multicolumn{2}{c?}{15s} & \multicolumn{2}{c}{50s} \\\hline
method & \sttt{KL} & \sttt{I/U} & \sttt{KL} & \sttt{I/U}  & \sttt{KL} & \sttt{I/U}
  \\\hline 

\sttt{FULL-PARALLEL [o]} &$-$ & 67.18 &$-$ & 67.70 & $-$ & 68.00 \\
\sttt{OURS-ADAM [o]} & $-$ & 66.45 & $-$ & 67.50 & $-$ & 68.07 \\
\hline\hline                                                                              
\sttt{FULL-PARALLEL} & \textbf{-3129799} & 67.21 & -3134437 & 67.72 & -3133010 & 68.01 \\
\sttt{ADHOC}         & -3129469          & 67.19 & -3134557 & 67.73 & -3136865 & 68.04 \\\hline
\sttt{OURS-FIXED}    & -3100079          & \textbf{67.76} & \textbf{-3135225} & \textbf{68.18} 
& -3138206 & 68.44 \\ 
\sttt{OURS-MOMENTUM}     & -3060405          & 66.20 & -3128121 & 67.39 & -3136543 & 68.18 \\
\sttt{OURS-ADAM}     & -3091787          & 67.08 & -3131624 & 68.02 & \textbf{-3138335} & \textbf{68.47} \\

\end{tabular}
\end{center}
\vspace{-0.2cm}
\caption{Results for semantic segmentation problem~\cite{Pascal-voc-2012} based on
  DeepLab-CRF~\cite{Chen15b}. For all the budgets, our method obtains better segmentation
  accuracy. Again, \texttt{FULL-PARALLEL} obtains lower KL faster, with a price of reduced
  performance. 
On the top, we provide results for the original DeepLab-CRF model without co-occurrence potentials (denoted by [o]), for which the KL divergence has therefore a different meaning and is not shown.}  
\label{tab:eval:segmentation-voc2012}
\vspace{-0.4cm}

\end{table*}

Quantitative results are given in Table~\ref{tab:eval:benchmark}. Our methods systematically
outperform the~\texttt{ADHOC} damping 
method. The~\texttt{SWEEP} method usually provides good performance, but is generally slow due
to its sequential nature. 

Fig.~\ref{fi:eval:benchmarks:steps} shows that our methods are less sensitive to
damping parameter changes than \texttt{ADHOC}.  
In Fig.~\ref{fi:eval:benchmarks:steps}, the vertical orange lines corresponds to the choice of the
damping parameter according to $d=L$, 
which can be computed directly by the power-method. Interstingly, for the
\texttt{GRID} dataset, which includes strong repulsive 
potentials, algorithms do not produce reasonable results when no damping is applied. On the other
hand, for the segmentation task, \texttt{SEG}, all the algorithms work well even without damping, in
accordance with the results of~\cite{Kraehenbuehl13} or Section~\ref{sec:fixedStep}. 

\vspace{-0.5cm}
\paragraph{Characters Inpainting}
Quantitative results in terms of average pixel accuracies and KL-divergences are given in
Table~\ref{tab:eval:dtf} and Fig.~\ref{fi:eval:convergence}~(a). Our method, especially when used with 
more advanced gradient descent schemes, outperforms all the baselines. \texttt{SWEEP} shows
relatively good performance, but does not scale as well in terms of the running time. See the bottom
row of Fig.~\ref{fig:intro:teaser} for an example of a result.

\vspace{-0.5cm}
\paragraph{People Detection}
Quantitative results, presented in Table~\ref{tab:eval:pom-soccer} and
Fig.~\ref{fi:eval:convergence}~(b), demonstrate that our method with a fixed step size,
\texttt{OURS-FIXED}, brings both faster convergence and better performance. Thanks to our
optimization scheme, the time required to get a Multiple Object Detection 
Accuracy(MODA,~\cite{Bernardin08}) within 3\% of the value at convergence is reduced by a factor of
two. This can be of big practical importance for surveillance applications of the
algorithm~\cite{BenShitrit14,Bagautdinov15}, in which it is required to run in
real-time. \texttt{SWEEP} exhibits much worse performance than our parallel method because 
of its greedy behaviour.  

\vspace{-0.5cm}
\paragraph{Semantic Segmentation}
Quantitative results are presented in Table~\ref{tab:eval:segmentation-voc2012} and
Fig.~\ref{fi:eval:convergence}~(c). We observe that a similar oscillation issue as noted
by~\cite{Vineet14} starts happening when the \texttt{FULL-PARALLEL} method is used in conjunction
with co-ocurrence potentials, producing even worse results than without those.  
Using our convergent inference method fixes oscillations and provides an
improvement of 0.5\% in the average Intersection over Union measure (\texttt{I/U}) compared to the
basic method without co-occurrence. This is a significant improvement that would be sufficient to
increase the position of an algorithm by 2 or 3 places in the official
ranking~\cite{Pascal-voc-2012}. What it represents is a big improvement in performance, as the ones
shown in Fig~\ref{fig:intro:teaser}, for at least 30-40 images out of total 1449. Note also,
that we obtain this improvement with minimal changes in the original code. By contrast,
authors~\cite{Chen15b} get similar or smaller improvements by significantly augmenting the training
set or by exploiting multi-scale features, which leads to additional computational burden.

%% file: discussion.tex
We have presented a principled and efficient way to do parallel mean-field inference in
discrete random fields. We have demonstrated that proximal gradient descent is a powerful
theoretical framework for mean-field inference, which unifies and sheds light on previous
approaches. Moreover, it naturally allows to incorporate existing adaptive  
gradient descent techniques, such as ADAM, to mean-field methods. As shown
in our experiments, it often brings dramatic improvements in performance. Additionally, we have
demonstrated, that our approach is less sensitive to the choice of parameters, which makes it
easier to use in practice.  

Importantly, our method brings considerable improvements in terms of performance on three realistic
Computer Vision problems, while being very easy to implement. 

Our method makes it possible to use mean-field inference with a much wider range of potential
functions, which was previously unachievable due to the lack of convergent optimization. Thus, there
is a large amount of possible future applications of our approach, especially in the tasks where
higher-order and repulsive potentials can bring additional benefits, not only in segmentation, but also in
object detection and localization.

%% file: top.bbl
\begin{thebibliography}{10}\itemsep=-1pt

\bibitem{Amari98}
S.-I. Amari.
\newblock {Natural Gradient Works Efficiently in Learning}.
\newblock {\em Neural computation}, 10(2):251--276, 1998.

\bibitem{Bagautdinov15}
T.~Bagautdinov, P.~Fua, and F.~Fleuret.
\newblock {Probability Occupancy Maps for Occluded Depth Images}.
\newblock In {\em Conference on Computer Vision and Pattern Recognition}, 2015.

\bibitem{BenShitrit14}
H.~BenShitrit, J.~Berclaz, F.~Fleuret, and P.~Fua.
\newblock {Multi-Commodity Network Flow for Tracking Multiple People}.
\newblock {\em IEEE Transactions on Pattern Analysis and Machine Intelligence},
  36(8):1614--1627, 2014.

\bibitem{Bernardin08}
K.~Bernardin and R.~Stiefelhagen.
\newblock {Evaluating Multiple Object Tracking Performance: the Clear Mot
  Metrics}.
\newblock {\em EURASIP Journal on Image and Video Processing}, 2008, 2008.

\bibitem{Bishop06}
C.~Bishop.
\newblock {\em {Pattern Recognition and Machine Learning}}.
\newblock Springer, 2006.

\bibitem{Boykov01b}
Y.~Boykov, O.~Veksler, and R.~Zabih.
\newblock Fast approximate energy minimization via graph cuts.
\newblock {\em Pattern Analysis and Machine Intelligence, IEEE Transactions
  on}, 23(11):1222--1239, 2001.

\bibitem{Campbell13}
N.~Campbell, K.~Subr, and J.~Kautz.
\newblock Fully-connected crfs with non-parametric pairwise potential.
\newblock In {\em Conference on Computer Vision and Pattern Recognition}, pages
  1658--1665, 2013.

\bibitem{Chen15b}
L.-C. Chen, G.~Papandreou, I.~Kokkinos, K.~Murphy, and A.~L. Yuille.
\newblock Semantic image segmentation with deep convolutional nets and fully
  connected crfs.
\newblock In {\em ICLR}, 2015.

\bibitem{DOrazio09}
T.~D'Orazio, M.~Leo, N.~Mosca, P.~Spagnolo, and P.~L. Mazzeo.
\newblock {A Semi-Automatic System for Ground Truth Generation of Soccer Video
  Sequences}.
\newblock In {\em International Conference on Advanced Video and Signal Based
  Surveillance}, pages 559--564, 2009.

\bibitem{Duchi10}
J.~C. Duchi, S.~Shalev-Shwartz, Y.~Singer, and A.~Tewari.
\newblock Composite objective mirror descent.
\newblock In {\em COLT}, pages 14--26. Citeseer, 2010.

\bibitem{Pascal-voc-2012}
M.~Everingham, L.~Van~Gool, C.~K.~I. Williams, J.~Winn, and A.~Zisserman.
\newblock The {PASCAL} {V}isual {O}bject {C}lasses {C}hallenge 2012 {(VOC2012)}
  {R}esults.
\newblock
  http://www.pascal-network.org/challenges/VOC/voc2012/workshop/index.html.

\bibitem{Fleuret08a}
F.~Fleuret, J.~Berclaz, R.~Lengagne, and P.~Fua.
\newblock {Multi-Camera People Tracking with a Probabilistic Occupancy Map}.
\newblock {\em IEEE Transactions on Pattern Analysis and Machine Intelligence},
  30(2):267--282, February 2008.

\bibitem{Frostig14}
R.~Frostig, S.~Wang, P.~Liang, and C.~Manning.
\newblock {Simple MAP Inference via Low-Rank Relaxations}.
\newblock In {\em Advances in Neural Information Processing Systems}, pages
  3077--3085, 2014.

\bibitem{Gelfand90}
A.~E. Gelfand and A.~F. Smith.
\newblock Sampling-based approaches to calculating marginal densities.
\newblock {\em Journal of the American statistical association},
  85(410):398--409, 1990.

\bibitem{Gorelick2014}
L.~Gorelick, Y.~Boykov, O.~Veksler, I.~Ben~Ayed, and A.~Delong.
\newblock Submodularization for binary pairwise energies.
\newblock In {\em Computer Vision and Pattern Recognition (CVPR), 2014 IEEE
  Conference on}, pages 1154--1161. IEEE, 2014.

\bibitem{Hoffman13}
M.~Hoffman, D.~Blei, C.~Wang, and J.~Paisley.
\newblock {Stochastic Variational Inference}.
\newblock {\em Journal of Machine Learning Research}, 14(1):1303--1347, May
  2013.

\bibitem{Emti15}
M.~E. Khan, P.~Baqu{\'{e}}, F.~Fleuret, and P.~Fua.
\newblock {Kullback-Leibler Proximal Variational Inference}.
\newblock In {\em NIPS}, 2015.

\bibitem{Kingma14}
D.~P. Kingma and J.~Ba.
\newblock Adam: {A} method for stochastic optimization.
\newblock {\em CoRR}, abs/1412.6980, 2014.

\bibitem{Koller09}
D.~Koller and N.~Friedman.
\newblock {\em {Probabilistc Graphical Models}}.
\newblock The MIT Press, 2009.

\bibitem{Kolmogorov15}
V.~Kolmogorov.
\newblock A new look at reweighted message passing.
\newblock {\em Pattern Analysis and Machine Intelligence, IEEE Transactions
  on}, 37(5):919--930, 2015.

\bibitem{Kraehenbuehl11}
P.~Kr{\"a}henb{\"u}hl and V.~Koltun.
\newblock {Efficient Inference in Fully Connected CRFs with Gaussian Edge
  Potentials}.
\newblock In {\em Advances in Neural Information Processing Systems}, 2011.

\bibitem{Kraehenbuehl13}
P.~Kr{\"a}henb{\"u}hl and V.~Koltun.
\newblock {Parameter Learning and Convergent Inference for Dense Random
  Fields}.
\newblock In {\em International Conference on Machine Learning}, pages
  513--521, 2013.

\bibitem{Minka01}
T.~P. Minka.
\newblock Expectation propagation for approximate bayesian inference.
\newblock In {\em Proceedings of the Seventeenth conference on Uncertainty in
  artificial intelligence}, pages 362--369. Morgan Kaufmann Publishers Inc.,
  2001.

\bibitem{Murphy99}
K.~Murphy, Y.~Weiss, and M.~Jordan.
\newblock {Loopy Belief Propagation for Approximate Inference: An Empirical
  Study}.
\newblock In {\em Onference on Uncertainty in Artificial Intelligence}, pages
  467--475, 1999.

\bibitem{Nowozin11}
S.~Nowozin, C.~Rother, S.~Bagon, T.~Sharp, B.~Yao, and P.~Kholi.
\newblock {Decision Tree Fields}.
\newblock In {\em International Conference on Computer Vision}, November 2011.

\bibitem{Polyak64}
B.~T. Polyak.
\newblock Some methods of speeding up the convergence of iteration methods.
\newblock {\em USSR Computational Mathematics and Mathematical Physics},
  4(5):1--17, 1964.

\bibitem{Saito12}
M.~Saito, T.~Okatani, and K.~Deguchi.
\newblock {Application of the Mean Field Methods to MRF Optimization in
  Computer Vision}.
\newblock In {\em Conference on Computer Vision and Pattern Recognition}, June
  2012.

\bibitem{Sun13}
X.~Sun, M.~Christoudias, V.~Lepetit, and P.~Fua.
\newblock {Real-Time {L}anding {P}lace {A}ssessment in {M}an-{M}ade
  {E}nvironments}.
\newblock {\em Machine Vision and Applications}, 2013.

\bibitem{Teboulle92}
M.~Teboulle.
\newblock Entropic proximal mappings with applications to nonlinear
  programming.
\newblock {\em Mathematics of Operations Research}, 17(3):pp. 670--690, 1992.

\bibitem{Vineet14}
V.~Vineet, J.~Warrell, and P.~Torr.
\newblock {Filter-Based Mean-Field Inference for Random Fields with
  Higher-Order Terms and Product Label-Spaces}.
\newblock {\em International Journal of Computer Vision}, 110(3):290--307,
  2014.

\bibitem{Wainwright08}
M.~Wainwright and M.~Jordan.
\newblock {Graphical Models, Exponential Families, and Variational Inference}.
\newblock {\em Foundations and Trends in Machine Learning}, 1(1-2):1--305, Jan.
  2008.

\bibitem{Winn2005}
J.~Winn and C.~Bishop.
\newblock {Variational Message Passing}.
\newblock In {\em Journal of Machine Learning Research}, pages 661--694, 2005.

\bibitem{Zheng15}
S.~Zheng, S.~Jayasumana, B.~Romera-Paredes, V.~Vineet, Z.~Su, D.~Du, C.~Huang,
  and P.~Torr.
\newblock Conditional random fields as recurrent neural networks.
\newblock In {\em International Conference on Computer Vision (ICCV)}, 2015.

\end{thebibliography}
